# Real-time Dexterous Telemanipulation with an End-Effect-Oriented Learning-based Approach

Haoyang Wang*, He Bai*, *Member, IEEE*, Xiaoli Zhang^, *Senior Member, IEEE*, Yunsik Jung^, Michel Bowman#, Lingfeng Tao*, *Member, IEEE*

*Abstract* — Dexterous telemanipulation is crucial in advancing human-robot systems, especially in tasks requiring precise and safe manipulation. However, it faces significant challenges due to the physical differences between human and robotic hands, the dynamic interaction with objects, and the indirect control and perception of the remote environment. Current approaches predominantly focus on mapping the human hand onto robotic counterparts to replicate motions, which exhibits a critical oversight: it often neglects the physical interaction with objects and relegates the interaction burden to the human to adapt and make laborious adjustments in response to the indirect and counter-intuitive observation of the remote environment. This work develops an End-Effects-Oriented Learning-based Dexterous Telemanipulation (EFOLD) framework to address telemanipulation tasks. EFOLD models telemanipulation as a Markov Game, introducing multiple end-effect features to interpret the human operator's commands during interaction with objects. These features are used by a Deep Reinforcement Learning policy to control the robot and reproduce such end effects. EFOLD was evaluated with real human subjects and two end-effect extraction methods for controlling a virtual Shadow Robot Hand in telemanipulation tasks. EFOLD achieved real-time control capability with low command following latency (delay<0.11s) and highly accurate tracking (MSE<0.084 rad).

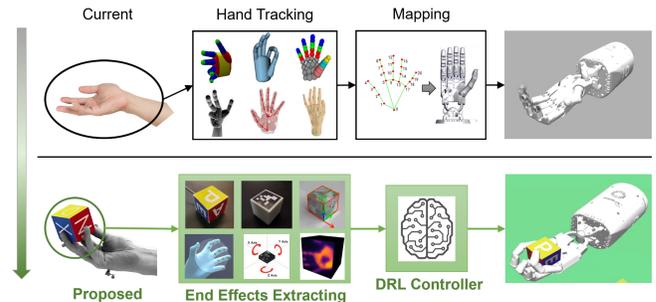

Figure 1. Compared with the current mapping-based telemanipulation, the End-Effects-Oriented Learning-Based Dexterous Telemanipulation (EFOLD) framework interpret human operator's interactive command with the objects to end effect features, then input to a Deep Reinforcement Learning policy that controls the robot to recreate such end effects.

## I. INTRODUCTION

Telemanipulation [1] is integral to advancing human-robot systems in contexts where contact-intensive and safety-critical manipulations are paramount, such as telesurgery, extraterrestrial exploration, and remote assembly and repair tasks. Compared with conventional teleoperation, telemanipulation involves a human operator continuously controlling a robot hand to apply dynamic and complex interaction with external objects. Thus, telemanipulation tasks exhibit significant challenges due to 1) the inherent physical disparity between human and robotic hands, characterized by differences in structure, size, and dexterity, 2) the complex physical interaction between the robot hand and the object that is difficult to be modeled and predicted, 3) the indirect control and perception of the remote environment to the human that cause delay, confusion and poor user experience.

Current telemanipulation methodologies predominantly focus on mapping the human hand onto robotic counterparts to replicate human hand motions through manually defined joint-to-joint [2] or simplified synergy-based mapping [3]. These approaches exhibit a critical oversight: they often neglect the physical interaction with objects, which is the priority of manipulation tasks. Consequently, the robot merely emulates the human hand, relegating the interaction with objects to the human's capacity to adapt and make laborious adjustments in response to the indirect and counter-intuitive observation of the remote environment, which not only diverts the operator's focus from the primary task but also impairs the overall task performance and potentially compromises the safety of the manipulation process. As a result, current approaches are neither sufficient to support real-time telemanipulation nor achieve good performance.

Deep reinforcement learning (DRL) [4] methods have demonstrated the ability to perform with high dexterity in the autonomous manipulation domain akin to human-like precision [5-7]. This approach excels in handling continuous and dynamic interactions with objects, offering adaptability, precise control, and robustness. Without dependence on predefined models, DRL learns from experience, enabling it to navigate complex dynamics and optimize manipulation strategies over time. The adaptability and scalability of DRL make it a potent solution for complex telemanipulation tasks. Despite these advantages, the application of DRL in telemanipulation faces significant challenges. A primary obstacle is translating human manipulation commands so a robot hand can comprehend and execute them. This translation involves capturing the nuances of human intent and conveying them so that the DRL system can process and act upon them effectively. Another challenge lies in training and deploying DRL-based controllers in real-world settings.

End-effect-oriented manipulation [8] prioritizes the outcomes or 'end effect' of interactions between a robot and the objects. This strategy focuses on understanding and controlling the physical consequences of manipulation, such as movement, tactile, force, and deformation, to achieve high precision and minimal error. For example, [8] studied force-based manipulation of deformable objects, the attempt

This work is supported by US NSF grant 2426469.

*H. Wang, H. Bai and L. Tao are with the Oklahoma State University, 563 Engineering North, Stillwater, OK, 74078, USA (e-mail: haoyangwang123@outlook.com, he.bai; lingfeng.tao@okstate.edu).

^Y. Jung and X. Zhang are with the Colorado School of Mines, Intelligent Robotics and Systems Lab, 1500 Illinois St, Golden, CO 80401, USA (e-mail: yunsikjung; xlzhang@mines.edu).

#M. Bowman is with the University of Pennsylvania, 3400 Civic Center Boulevard Building 421, Philadelphia, PA 19104, USA (e-mail: michael.bowman@pennmedicine.upenn.edu).

in [9] teaches the robot to grasp objects with the thermal images of human demonstrations. End effects are crucial in tasks that require delicate handling and interactions with objects of varying fragility, size, and shape. Such advantages make end effects perfect for bridging the gap between human command and DRL policy in task-based telemanipulation.

This work develops the **End-Effects-Oriented Learning-Based Dexterous Telemanipulation (EFOLD)** framework (Fig. 1), which fundamentally redefines the roles of human operators and robotic systems in telemanipulation to release the human operator's burden of control. The end effects are defined as the information that describes the physical consequences of manipulation, such as movement, tactile, force, friction, and deformation. EFOLD prioritizes recreating the end effects of interactions between a robot and the objects in a learning-based control manner. EFOLD models telemanipulation as a Markov Game [10] where the robot and the human are considered agents. We propose to use the end effect as the intermediary to interpret the human operation to robot understandable command that covers the physical interaction information. With this modeling process, the robot control policy can be solved by training a DRL agent, which uses the end effect as the manipulation goal. The key innovation of this part is that we utilize the offline training and online evaluation strategy, which trains a generalizable DRL policy to track end-effect goals that are randomly generated but follow specific rules to mimic human behaviors. We assume that a well-trained generalizable end-effect-oriented DRL policy covers the end-effect state space that the human operator provides. In such a strategy, the human operator only needs to join during testing.

The contributions of this work are summarized as follows:
1) Model the telemanipulation task as a Markov Game to provide the mathematical foundation for the DRL-based telemanipulation framework.
2) Categorize end effect extraction methods internally and externally and analyze their practicability.
3) Develop human-offline training and human-online testing strategies to free up human involvement in training to save time and improve cost-effectiveness.
4) Evaluate the EFOLD framework with real human subjects in telemanipulation tasks using a virtual Shadow Dexterous hand and test the manipulated object as a joystick to play a game.

## II. RELATED WORK

### A. Mapping-based Telemanipulation

Traditional mapping focuses on grasping tasks by projecting the human hand pose to the robot [2]. Inverse kinematics determines the empirical mapping of which human joints control which robot joints [11]. This style of approach works for less sophisticated robots. Recently, synergy-based mapping has been studied through joint space [3], Cartesian fingertip space [12], and virtual object [24], which further simplifies the mapping space. However, the grasp may look unnatural and lead to incorrect forces applied to the object.

Recent advancement in end-to-end mapping enables high-performance human hand tracking. TeachNet [13] uses shadow images and an autoencoder to control the robot to recreate human gestures. DexPilot [2] uses point clouds and deep neural networks to process the RGBD image of the human hand to generate human hand configurations in real-time. However, the ignorance of physical interaction still puts the control burden on the human and significantly reduces the manipulation speed (4x slower). Regardless of the mapping strategy, the robot replicates and imitates the static poses a human commonly generates. These forms of telemanipulation essentially identify a human gesture and formulate a corresponding robotic gesture and do not consider task dynamics and physical interaction with the external object.

### B. DRL-based Teleoperation

DRL-based methods are widely studied in general teleoperation topics. Shared autonomy [14] models human behavior as a Markov Decision Process (MDP) [15] and focuses on the approaching/navigation phase of the task for grasping tasks. In [14], a DRL-controlled robot assists the user toward a goal position by predicting the user intent and integrating their inputs. Recently, [16] proposed a human-in-the-loop RL to achieve both following user commands and deviating from the user's actions when they are suboptimal for the Lunar Lander game. Yet, these methods are for navigating the robot toward a target position, so-called target approaching, but not for in-hand object manipulation. Due to the physical discrepancy of hand structures and interaction between object and hand, DRL for target approaching is unsuitable for in-hand object telemanipulation.

### C. End-Effect-Oriented Autonomous Manipulation

End effects are studied in autonomous manipulation tasks emphasizing the importance of the outcomes or 'end effects' of the interactions with the objects. Deep learning models [17] have recently been developed to predict the spatial end effects. ContactDB [9] uses thermal images to train the robot to learn grasping from human demonstrations. Soft object manipulation also uses tactile information such as contact points, force, and pressure [18]. Higher-level end effects such as grasp quality [19] and affordance [20] are studied based on contact points, coverage area, and distance to the object center. These studies demonstrated the effectiveness and promise of an end-effect-oriented strategy for autonomous manipulation. However, how this can be used for in-hand object telemanipulation where human inputs need to be considered in the loop has rarely been studied.

## III. END-EFFECTS-ORIENTED LEARNING-BASED DEXTEROUS TELEMANIPULATION FRAMEWORK

This section introduces the modeling of the DRL-based telemanipulation in III.A. The categorization of the potential end effect features is discussed in III.B. The offline training and online testing strategy is presented in III.C.

### A. Multi-agent Modeling and Representation of DRL-based telemanipulation

DRL methods have demonstrated their capability to handle dexterous in-hand manipulation tasks such as rotating a block to a goal pose [5] or solving a Rubik's cube [6]. We start with the mathematical modeling of the DRL-based telemanipulation approach. Typically, a single-agent DRL problem is modeled as a Markov Decision Process (MDP), which is defined as a tuple $\{S, A, R, \gamma\}$, where $S$ is the state space of the environment, $A$ is the set of available actions, $R: S \times A \to \mathbb{R}$ is the reward that is returned by the environment, and $\gamma \in [0,1]$ is the discount factor. The purpose of DRL training is to maximize the reward during the task. Unlike single-agent tasks, telemanipulation tasks involve two agents: **a human operator** and **a robot**. Thus, we model the telemanipulation problem as a Markov Game, an extension of MDP with multiple agents. Specifically, a telemanipulation task contains two MDPs shown in Fig. 2. Each agent's behavior follows an MDP and has its policy: the human operator as an agent, who is interacting with the object with state transition denoted as $\{H_t \rightarrow H_{t+n}\}$, and an autonomous agent following the human's command to manipulate the object. The Markov game is a tuple $\{H, S, A, R, \gamma\}$. In the telemanipulation task, the robot is following the human, so its policy $\pi_\beta: S \times H \times R$, which depends on its state, the human command, and the operation reward. However, in the real world, the human mind is a black box, making the human MDP a Partially Observable MDP (POMDP) [15], where the transition, action, and reward cannot be directly assessed, and only the human's physical interaction with the object can be observed. The Markov Game model leads us to the innovation of the end-effect-oriented approach to transform partial observation into robot-understandable end-effect goals and helps clear the relationship between humans and robots to develop DRL algorithms and training strategies.

*B. End Effect Feature Categorization and Applicability*

The Markov Game modeling reveals the necessity for an interpretation layer to bridge Human POMDP and Robot MDP. In this section, we utilize the end effects as the interpretative link to describe how the human operator interacts with the object. To ensure the viability of our approach in real-world complex 3D space operation, we delve into the categorization of end effect features.

In our telemanipulation setup, humans need to manipulate a demonstrating object. The end effect information is extracted from the sensors by tracking the object. We first identify several potential sensory technologies from the literature that can be used for end effect extraction for telemanipulation, including IMU, accelerometer, gyroscope, tactile, pressure, vibration, RGB/RGBD camera, data glove, IR Sensor, laser, MM wave, and thermal image. Based on how the end effect features are extracted, they are divided into two categories: **Internal extraction**, which means that the sensors can be integrated into the demonstrating object, and **External extraction**, where the sensors are external devices set up around the environment (Table I). This categorization aids in selecting features for specific tasks involving confined spaces, extra precision, and multi-model perceptions. We further analyze the practicability of the categorized sensors based on the following end-effect features:

**Translational features.** Translational information tells the robot how humans want to place the object. The identified translational end effect features in 3D space including but not limited to the position, denoted as $(x, y, z)$, the speed, denoted as $(\dot{x}, \dot{y}, \dot{z})$, and acceleration, denoted as $(\ddot{x}, \ddot{y}, \ddot{z})$. The internal sensors that extract these features include accelerometers and gyroscopes. With Computer Vision [17], nearly all vision-based external sensors can be used, including RGB/RGBD cameras, reflective trackers, IR sensors, and Laser tracking.

TABLE I. END EFFECT FEATURES CATEGORIZATION

| Internal | External |
|---|---|
| IMU | RGB/RGBD |
| Accelerometer | Data glove |
| Gyroscope | IR Sensor |
| Tactile | Laser |
| Pressure | MM wave |
| Vibration | Thermal |

**Rotational features.** Rotational information tells the robot how humans want to rotate the object. The identified rotational end effect features in 3D space including but not limited to the rotational position, denoted as $(\psi, \theta, \varphi)$, the rotational speed, denoted as $(\dot{\psi}, \dot{\theta}, \dot{\varphi})$, and rotational acceleration, denoted as $(\ddot{\psi}, \ddot{\theta}, \ddot{\varphi})$. The internal sensors include IMU, accelerometers, and gyroscopes. Like translational features, nearly all vision-based external sensors can be used.

**Tactile features.** Tactile information [21] can help the robot to understand further the dexterous interaction between the specific hand components (like fingertips and palm area) and the object. It potentially constrains the DRL policy's behavior and drives the robot hand to behave like the human hand. The identified tactile features including but not limited to the touch position, denoted as $(\tau_x, \tau_y, \tau_z)$, touch force vector, denoted as $\vec{F}(x, y, z)$, and touch pressure, denoted as $P$. The internal sensors that extract these features include tactile, pressure, deformation, and skin sensors. External sensors include data gloves that the human operator can wear.

*C. Human-Offline Training and Human-Online Testing*

Conventional training of autonomous DRL policy can usually be accelerated with fast simulation and parallel training on a supercomputer. However, the biggest challenge of adopting DRL methods to human-robot systems is that the training of DRL policy needs human involvement, and humans can only operate in real-time, which brings high costs. The key innovation of this part is that we utilize the human-offline training and human-online testing strategy (Fig. 3), which trains a generalizable DRL policy to track an end-effect's goal trajectory that is randomly generated to mimic human behaviors. A well-trained, generalizable, end-effect-oriented DRL policy can cover the goal state space from the human operator. In such a strategy, the human operator only needs to join during testing.

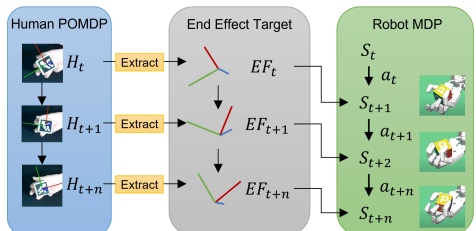

Fig. 2. We model the telemanipulation tasks as a Markov Game that involves two agents: a human operator and a robot. The robot's MDP depends on the operation command from the state of human's POMDP because the human's mind is a black box and only the end effects that the human applied to the object can be observed by the robot.

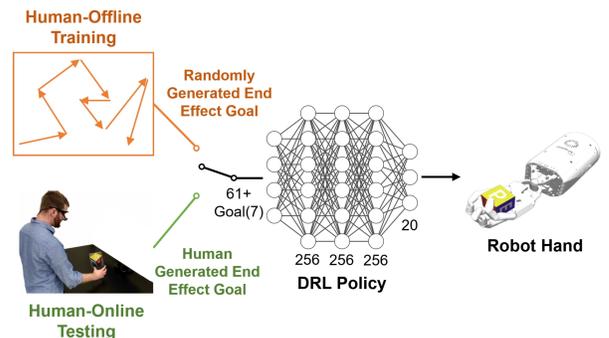

Fig. 3. Illustration of the human-offline training and human-online testing strategies. During training, the human operator is offline, the DRL policy is trained with randomly generated end effect goal. After training, the DRL policy will be generalizable to follow the target provided by real human operators. Then the human operator can control the robot in testing.

## IV. EXPERIMENTS DESIGN AND EVALUATION METRICS

### A. Task Design

The EFOLD framework requires a capable test platform to derive viable dexterous telemanipulation tasks. The framework will be evaluated in a simulated environment for easy training and testing. Specifically, we adopt the Shadow hand environments from the OpenAI GYM Robotics platform [5], which runs on the MuJoCo physics simulator [22]. The telemanipulation scenario is designed to control the robot hand to rotate a block placed on the hand's palm with a random initial pose (Fig 4). The task is manipulating the block around the Z-axis to achieve the goal pose. Two reward functions are designed to compare the performance of the DRL policy under different rewards:

**Sparse:** the reward function gives a binary reward of $r_b = 0$ if the goal has been achieved and $r_b = -1$ if the task failed. The sparse reward only evaluates if the goal is achieved at the last time step, giving the policy more freedom to find ways to achieve the end-effect goal.

**Dense:** the reward function is defined as:
$$R = -2 \cdot arccos(|q_a \overline{q_g}|) + r_b \quad (1)$$
where $q_a$ is the object angle while $q_g$ is the goal, both in quaternion form. $arccos$ function calculates the minimal angular distance between the current and goal positions. The dense reward function provides a continuous reward signal during the manipulation process, providing more guidance to the target, but is harder to train. Thus, we set the criterion that a goal is achieved if the angular distance is less than 0.1 rad for the Sparse and 0.05 rad for the Dense.

The agents are running at a time step of 0.04s. The PC hardware for training includes an Intel 12900K, a Nvidia RTX3080ti, and 64 GB of RAM. The Deep Deterministic Policy Gradient (DDPG) algorithm trains the DRL agent. The Hindsight Experient Replay method is used to improve the sample efficiency. A 5-layer fully connected neural network is designed as the policy network (Fig. 3). The inputs consist of 24 robot joint angular positions, and 24 angular velocities, 13 object states comprising Cartesian position and rotation represented by a quaternion, along with its linear and angular velocities, and an end effector goal in Cartesian position and rotation represented by a quaternion, totaling 68 inputs. The output is 20 joint actuators of the hand. Most hyperparameters are from [24], but with changes to the total epoch to 200.

In the block telemanipulation task, the end effect feature is defined as the block's rotation angle $\varphi$ around the Z-axis. Due to limited resources and implementation difficulties in real-world sensory setup, we selected IMU as the representative sensor from the internal category and RGB camera as the external category. A wireless IMU is placed inside a 3D-printed block to read the rotation angle. A top-down camera is used to track an Aruco code [23] that is glued to the top of the block (shown in Fig. 4) for ease of set up. To the best of the author's knowledge, there are no end-effect-based telemanipulation methods that can be used as a baseline comparison. The only comparable approach is a mapping method, DexPilot [2]. However, directly comparing the performance is inappropriate because of the significant difference in experiment setup and non-real-time capability (about 4x slower than our method). An extension test is designed to evaluate EFOLD's real-world performance. The telemanipulated object is treated as a joystick to play the game Pong. The game points and number of failures of direct control by the human and telemanipulated control by the object are recorded to assess the performance.

### B. Evaluation Metrics and Quantitative Testing

During the human-offline training, the goal positions are randomly generated within the range of $(-\pi, \pi)$ $rad$. Instead of logging the episode reward and average reward, the task success rate of the policy is recorded at the end of each epoch for a direct and unbiased comparison. The success rate is the percentage of successful cases in a validation set with 50 trials. Each trial randomly generates an initial block position and sequential end-effect goals. Each dense and sparse configuration is trained 5 times with 200 epochs to obtain statistical results. Because human commands are difficult to repeat, we designed the following two-goal trajectories from traditional control theory to acquire quantitative data for performance evaluation:

**Sinusoidal goal:** The rotational end effect trajectory is designed to follow the following Sinusoidal function:
$$G_s = \alpha sin(\omega x)$$
where $\alpha = (0.5, 1)$ is the amplitude, and $\omega = (0.1, 0.05)$ is the frequency. The Sinusoidal goal helps to measure the performance under periodic commands. The following metrics are designed for the Sinusoidal trajectory:

<u>MSE:</u> The mean square error is calculated as:
$$MSE = \frac{1}{n}\sum_{i=1}^{n}(\varphi_i - G_i)^2 \quad (2)$$
where $n$ is the number of steps, $\varphi_i$ is the object rotation angle around the Z-axis, $G_i$ is the end-effect goal. MSE evaluates the overall following performance during telemanipulation.

<u>Average Latency:</u> The average latency is calculated as
$$L = \theta_a - \theta_g \quad (3)$$
where $\theta_a$ is the phase of the actual trajectory of the object, $\theta_g$ is the phase of the end-effect goal trajectory, which are obtained from the dominant frequency after Fourier transform. The average latency evaluates the manipulation delay, a critical measure to show real-time control capability.

<u>Saturation:</u> The proportion of instances where the robot executes maximum magnitude movements to the total executions is denoted as $Sat$. Saturation quantifies the extent of exaggeration in the robot's movements, offering a comprehensive evaluation of manipulation performance.

<u>Average Energy Consumption:</u> The average energy consumption is calculated as:
$$\bar{E} = \sum_{i=1}^{n}\sum_{j=1}^{m} a_{ij} \quad (4)$$

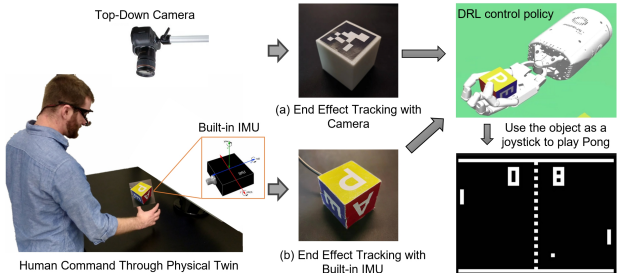

Fig. 4. The experiment includes a human remotely controls a dexterous robot hand to rotate a cube around the Z-axis. We used two extraction methods: (a) a top-down camera and (b) a built-in IMU. The extracted end effect goal will be used to control the robot hand. To evaluate the applicability, the recreated object is used as a joystick to play Pong.

where $n$ is the number of steps, $m$ is the number of Shadow hand joints, $a_{ij}$ is the magnitude of motion for the j[th] joint at the i[th] step. This measure enhances our understanding of managing and optimizing energy consumption in telemanipulation to improve energy efficiency.

**Step goal:** The end-effect goal is designed as a step signal that sets the object's initial rotation to 0 and the end-effect goal to $G_{step} = 1\ rad$ at the 50[th] step. The step response helps to evaluate the telemanipulation performance for commands that change in a short time. The following metrics are designed for the step trajectory based on the traditional control theory:

*Steady State Error:* This measure is calculated as
$$e_{ss} = \lim_{i \to n} e(i) \times 100\% \quad (5)$$
where error $e(i) = \varphi_i - G_{step}$. It evaluates how accurately the DRL policy can track the end-effect goal.

*Overshoot:* We calculate the max-percent overshoot:
$$OS = \frac{\varphi_p - G_{step}}{G_{step}} \times 100\% \quad (6)$$
where $\varphi_p$ is the peak object rotational angle. The overshoot measures the aggressiveness of the DRL control policy.

*Settling Time:* The time $t_{st}$ took for the robot to reach and stay within a range of 5% of the final goal. It measures how fast the robot stabilizes after being subjected to perturbation.

*Peak Time:* The time required for the response to reach the peak value for the first time, which is denoted by $t_p$. It indicates how quickly the control policy can react to the human command. For all measures, the lower is better.

After the quantitative evaluation, the human operators will join the operation of the DRL policy to control the Shadow hand for the block manipulation task. We record the tracking performance in temporal space to show the EFOLD's real-time tracking capability, and the tracking MSE will be calculated to show the overall performance. The video of the testing is included in the supplementary materials.

## V. RESULTS AND DISCUSSION

### A. Human-Offline Training

The results of the training process are shown in Fig. 5. The policy performance increases at a similar speed in early training for both sparse and dense rewards. On average, sparse achieved a slightly higher success rate $\approx 0.8$ than dense success rate $\approx 0.7$ when the training finished. This is because the dense reward constrains the whole manipulation process with a stricter success tolerance (0.05 rad) than Sparse, which makes it more difficult to find an optimal solution in the optimization view. In the next section, the sophisticated performance evaluation metrics will help us to identify which policy is more suitable for telemanipulation.

### B. Quantitative Testing

The quantitative testing results for the sinusoidal goal tracking are shown in Table II. Overall, the policies from dense and sparse rewards can keep a good track of the end-effect goal with the highest latency of 2.89 steps (each step is 0.04s) for Dense, which is around **0.11s** on a real-world time scale. More visualized results can be found in the supplementary video. As a comparison, Dexpilot [2] needs to accelerate their demo 4 times to achieve similar results as ours. The tracking performance deteriorates as the sinusoidal amplitude increases and improves as the sinusoidal frequency decreases, owing to the alteration in task complexity. Specifically, the policy trained with dense rewards (Dense) has higher latency, lower saturation, and higher energy consumption than that trained with sparse rewards (*Sparse*). This is because *Dense* is trained with continuous guidance during the manipulation process, which teaches to follow the goal trajectory precisely, so the saturation level is low. However, such behavior slows the manipulation with higher delay and increases energy consumption.

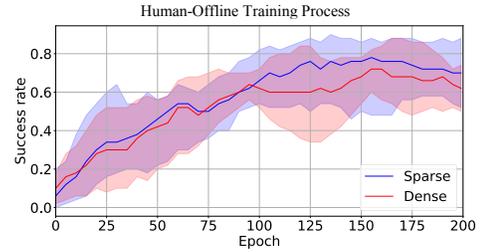

Fig. 5. The training process of dense and reward.

TABLE II. SINUSOIDAL GOAL TRACKING PERFORMANCE (200 TESTS)

|  | $\alpha$ | $\omega$ | $MSE(rad)$ | $L(step)$ | $Sat(\%)$ | $\bar{E}$ |
|---|---|---|---|---|---|---|
| Dense | 0.50 | 0.10 | 0.01 | -1.93 | 6.75 | 666.92 |
|  | 0.50 | 0.05 | 0.01 | -1.84 | 4.43 | 518.58 |
|  | 1.00 | 0.10 | 0.32 | -2.89 | 23.69 | 720.02 |
|  | 1.00 | 0.05 | 0.03 | -2.43 | 10.71 | 627.33 |
| Sparse | 0.50 | 0.10 | 0.02 | -1.61 | 20.59 | 242.51 |
|  | 0.50 | 0.05 | 0.01 | -1.80 | 10.08 | 132.60 |
|  | 1.00 | 0.10 | 0.26 | -0.28 | 19.99 | 499.79 |
|  | 1.00 | 0.05 | 0.08 | -0.46 | 18.03 | 349.01 |

The results for the step goal tracking are shown in Table III. *Dense* and *Sparse* have similar peak times (153 and 163) and settling times (102 and 109) because these two measures depend more on the dynamic systems, which are the same in this testing. However, *Dense* achieved a much lower overshoot and steady-state error than *Sparse*, which means better performance. This corresponds to the training difficulty explained in the sinusoidal tracking results.

Fig. 6 shows the best-performing sinusoidal and step goal-tracking tests. *Dense* (MSE = 0.004 rad) outperforms *Sparse* (MSE = 0.009 rad) in the sinusoidal goal but vice versa in the step tracking. The results reveal the policy behavior difference, where the *Dense* performs better with smooth and continuously changing goals. *Sparse* performs better with fast-changing goals. This is because *Sparse* reacts faster than *Dense,* but it is also difficult to reduce the tracking error.

### C. Human-Online Testing

In the human-online testing, a human operator conducted 5 tests in the block telemanipulation tasks with *Dense* and *Sparse,* which have the best quantitative performance. The results are shown in Table IV. Telemanipulation section. As explained in section IV, we only calculated the MSE and Sat

TABLE III. STEP GOAL TRACKING PERFORMANCE (200 TESTS)

|  | $t_p(step)$ | $t_{st}(step)$ | $OS(\%)$ | $e_{ss}(\%)$ |
|---|---|---|---|---|
| Dense | 153 | 102 | -0.57 | -0.87 |
| Sparse | 163 | 109 | -1.43 | -1.51 |

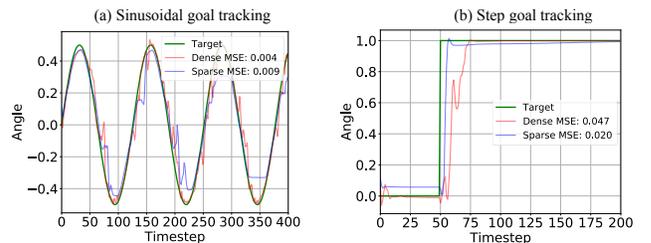

Fig. 6. Tracking performance of (a) Sinusoidal Goal and (b) Step Goal

TABLE IV. REAL-WORLD TESTING IN A TELE-PLAYED GAME* (5 TESTS)

| | | MSE(rad) | Sat | Drop | Hit | Failure |
|---|---|---|---|---|---|---|
| | | Telemanipulation | | | Play Pong | |
| Sparse | IMU | 1.75 | 20.71% | 60% | 11 | 2 |
| | Camera | 0.25 | 17.75% | 40% | 13 | 7 |
| Dense | IMU | 0.07 | 2.25% | 20% | 20 | 3 |
| | Camera | 0.03 | 1.60% | 0 | 23 | 2 |
| Direct | IMU | - | - | - | 25 | 0 |
| | Camera | - | - | - | 25 | 0 |

*The total numbers of hit and failure are different because the robot drops the block, causing a failure and end in the game early.

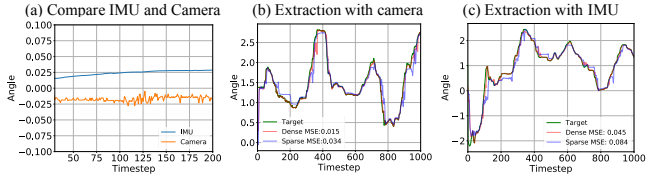

Fig. 7. Human-online testing: (a) extraction with IMU, (b) extraction with camera, (c) compare IMU and camera extraction performance.

measures due to the inconsistency of human behaviors. Considering reward, *Dense* performs better than *Sparse* with an MSE smaller than 0.07 rad and a saturation level 0.02%. The percentage of tests when the object drops out of the hand is calculated, denoted as *drop*. *Dense* achieved a lower drop rate than *Sparse*. This is because *Dense* is more stable than *Sparse*, as shown in Fig. 6. Considering sensors, the IMU performs worse than the camera with much higher MSEs and saturation levels. To find the reason, we recorded the telemanipulation test for the IMU and camera extraction methods in the best-performed human-online tests. Fig. 7(a) first shows the comparison of IMU and camera tracking performance when tracking a static object. The IMU has a clear signal but drifted over time, which requires frequent calibration during the experiment. On the contrary, the camera tracking has higher noise level but also shows stable and accurate tracking. For EFOLD telemanipulation, end-effect tracking accuracy is the most critical factor for the DRL policy to apply correct control. Thus, the IMU performs worse than the camera, but still has acceptable performance. In Fig. 7(b, c), even the lower-performance IMU can telemanipulate in real-time, with a the 0.08 rad MSE, which proves the real-time and tracking performance of our EFOLD framework.

The results of Pong are shown in Table IV Play Pong section. Overall, *Dense* achieved a higher number of successful hits of the ball (43) than *Sparse* (24), showing consistent advantage. Camera outperforms IMU with a higher number of successful hits (36) than IMU (31) because IMU's low performance caused 60% in sparse and 20% in dense object drop rates, higher than the camera (40% in sparse and 0 in dense), result in more failures than the camera.

In summary, the results show that dense reward is a better option for EFOLD because the trained DRL policy has a more stable manipulation behavior. But the internal and external feature extraction methods can be applied to the EFOLD framework in real-world applications. However, the sensor performance will affect the overall system performance. Our future work will extend the EFOLD framework to real-world experiments with physical robot hands and study the selection and optimization of sensors with multi-model perception for safe operation and intelligent feedback for humans.